\documentclass[runningheads]{llncs}
\usepackage[T1]{fontenc}
\usepackage{graphicx,verbatim}
\usepackage{multirow}
\usepackage{graphicx}
\usepackage{amsmath}
\usepackage{booktabs}
\usepackage{makecell}
\usepackage{hyperref}
\usepackage[inkscapelatex=false]{svg}
\begin{document}
\title{Dynamic Gradient Sparsification Training for Few-Shot Fine-tuning of CT Lymph Node Segmentation Foundation Model}
\titlerunning{Few-shot Fine-tuning for LN-Seg Foundation model}

\author{Zihao Luo\inst{1,2} \and Zijun Gao\inst{3} \and Wenjun Liao\inst{1} \and Shichuan Zhang\inst{1} \and Guotai Wang\inst{2,4} \and Xiangde Luo\inst{5}}

\authorrunning{Z. Luo et al.}
\institute{
$^1$Department of Radiation Oncology, Sichuan Cancer Hospital and Institute, University of Electronic Science and Technology of China, Chengdu\\
$^2$School of Mechanical and Electrical Engineering, University of Electronic Science and Technology of China, Chengdu, China\\
$^3$Department of Computer Science and Engineering, The Chinese University of Hong Kong, Sha Tin, Hong Kong.\\
$^4$Shanghai AI Laboratory, Shanghai, China.\\
$^5$Department of Radiation Oncology, Stanford University School of Medicine, Palo Alto, California\\
\email{luoxd96@stanford.edu}\\
}

\maketitle              

\begin{abstract}
Accurate lymph node (LN) segmentation is critical in radiotherapy treatment and prognosis analysis, but is limited by the need for large annotated datasets. While deep learning-based segmentation foundation models show potential in developing high-performing models with fewer samples, their medical adaptation faces LN domain-specific prior deficiencies and inefficient few-shot fine-tuning for complex clinical practices, highlighting the necessity of an LN segmentation foundation model. In this work, we annotated 36,106 visible LNs from 3,346 publicly available head-and-neck CT scans to establish a robust LN segmentation model (nnUNetv2). Building on this, we propose \textbf{D}ynamic \textbf{G}radient \textbf{S}parsification \textbf{T}raining (DGST), a few-shot fine-tuning approach that preserves foundational knowledge while dynamically updating the most critical parameters of the LN segmentation model with few annotations. We validate it on two publicly available LN segmentation datasets: SegRap2023 and LNQ2023. The results show that DGST outperforms existing few-shot fine-tuning methods, achieving satisfactory performance with limited labeled data. We release the dataset, models and all implementations to facilitate relevant research: \url{https://github.com/Zihaoluoh/LN-Seg-FM}.


\end{abstract}
\section{Introduction}
Accurate lymph node (LN) segmentation in computed tomography (CT) scan is critical for radiotherapy planning, prognosis and follow-up analysis~\cite{ji2023lymph}, yet manual delineation of all visible LN remains clinically impractical due to time constraints. While deep learning methods have advanced automated segmentation~\cite{li2022npcnet,yu2024effective}, their deployment is hindered by demanding data needs stemming from LNs' anatomical heterogeneity and inter-patient variability. Emerging foundation models address data scarcity challenges~\cite{zhang2024challenges,moor2023foundation} but face two critical limitations in LN-specific applications: First, generalist  ~\cite{kirillov2023segment} or non-LN-specific models~\cite{Liu_2023_ICCV,chen2024ma,ma2024segment,ye2023uniseg} lack domain-specific priors for subtle LN boundary characterization in medical images, leading to suboptimal performance. Second, their excessive computational complexity impedes clinical translation to resource-limited settings. These gaps highlight the unmet need for LN-specialized foundation models in the reality of hospitals that simultaneously achieve:  LN variation-robust generalization ensuring exposure to diverse LN variations, computational efficiency to operate within hardware constraints, and stable retraining protocols which is a critical requirement given the constant influx of new patient cohorts with evolving disease patterns in clinical practice. These requirements are interdependent—a prerequisite for practical adoption in radiotherapy workflows.

To address the aforementioned requirements, we conducted an exploratory study focusing on developing and deploying the LN segmentation foundation model. In the first phase, we annotated over 36,106 visible LNs in 3,346 CT scans from a publicly available head and neck (HN) cancer cohort~\cite{welch2024radcure}, capturing the variability in LN morphology and distribution to establish a solid training base. Considering both performance and efficiency, we selected nnUNet, the most widely used backbone, for its strong capabilities~\cite{isensee2021nnu,isensee2024nnu}. For the retraining for model updating or downstream deployment, parameter-efficient fine-tuning (PEFT) is considered due to its advantage of not requiring additional model structures or adjustments to inference settings~\cite{liu2022few,mosbach2023few}. However, current methods typically fixed tuning on a subset of model parameters or added new structures to maintain model stability~\cite{hulora,zaken2022bitfit,frankletraining}, limiting their ability to capture the complex individual heterogeneity in medical scenarios, particularly in terms of cohort and disease variations, hindering knowledge transfer across large gaps. Therefore, a few-shot fine-tuning method that dynamically balances model stability and flexibility is urgently needed.

To bridge these gaps, we introduce a novel approach called \textbf{D}ynamic \textbf{G}radient \textbf{S}parsification \textbf{T}raining (DGST). Compared with previous work~\cite{zhang2024gradient,hulora}, it not only maintains stability through sparse parameter updates but also improves efficiency and performance by implementing a dynamic sparsification process via gradients. Specifically, during the fine-tuning phase of the foundation model, DGST dynamically selects and updates the most critical parameters based on the gradient at each iteration. This dynamic selection process ensures a balance between model stability and plasticity. We investigate the LN segmentation foundation models' transferability and DGST's effectiveness on two representative downstream tasks: one within the same anatomical region (SegRap2023) and one across different regions (LNQ2023) to evaluate DGST's ability to handle complex variations in medical scenarios. The contributions of this work can be summarized as follows: (1) This pioneering exploratory study on foundation models for LN segmentation involves the annotation and public release of 36,106 visible LNs in 3,346  CT scans from a publicly available HN cancer cohort~\cite{welch2024radcure}, providing a valuable basis for future research in this area. (2) We propose DGST, a few-shot fine-tuning method tailored to balance the stability and flexibility of foundation models in medical scenarios involving complex disease progression and LN anatomical variability. (3) Comprehensive validation on HN LN segmentation using the SegRap2023 dataset and mediastinal LN segmentation with the LNQ2023 dataset demonstrates the superior performance of DGST over existing methods. Besides, our DGST method achieves comparable performance to sufficient data setting in SegRap2023 with only 10 annotated samples and significantly narrows the performance gap in LNQ2023 with 20 annotated samples.
\section{Method}
\begin{figure}[t]
\includegraphics[width=\textwidth]{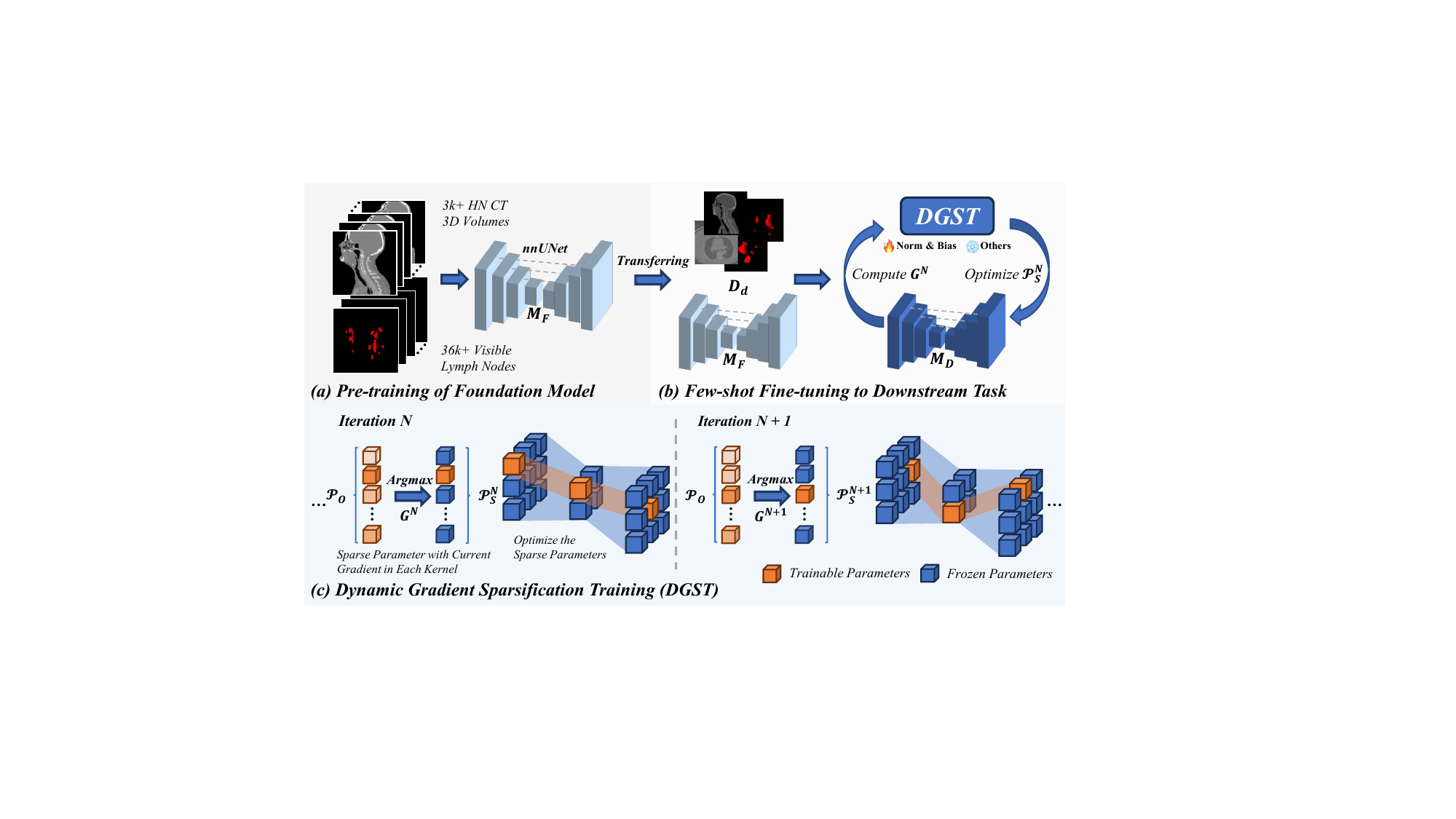}
\caption{(a) Pre-training of the foundation model using 3k+ HN CT volumes and 36k+ visible lymph node annotations. (b) Few-shot fine-tuning to downstream tasks by transferring the pre-trained model to new datasets via Dynamic Gradient Sparsification Training (DGST). (c) DGST methodology: At each iteration, the parameters $\mathcal{P}_O$ are sparsified to $\mathcal{P}^N_S$ using the current gradient $G^N$ for each kernel, and then optimized.} \label{fig1}
\end{figure}
We consider a scenario in which a pre-trained LN segmentation foundation model $M_F$ is adapted to downstream tasks with limited annotated samples $\mathcal{D}_d$. An overview of the framework is shown in Fig. \ref{fig1}, and the subsequent sections formally introduce each of its components.
\subsection{Pre-training of Foundation Model}
To comprehensively capture the variability of LNs, we utilized the large-scale HN CT dataset RADCURE~\cite{welch2024radcure}, denoted as $\mathcal{D}_F$. We re-delineated this dataset with all visible LNs to encounter the advanced radiotherapy treatment trends~\cite{liao2023visible}, and to employ a fully supervised manner to construct the LN segmentation foundation model. The pre-training of the foundation model is represented as:
\begin{equation}
    M_F=\arg\min_{\theta_F\in M_F} \mathcal{L}_{CE+Dice}(\mathcal{D}_F:\theta_F)
    \label{foundation}
\end{equation}
Where $M_F$ and $\theta_F$ represent the foundation model and its parameters, respectively; $\mathcal{L}_{CE+Dice}$ denotes the combined cross-entropy and Dice loss.
\subsection{Few-shot Fine-tuning to Downstream Task}
Few-shot fine-tuning is considered an efficient approach for knowledge transferring from foundation models, as it does not require modifications to the pre-training process and facilitates easier deployment than other methods~\cite{liu2022few,mosbach2023few}. In the few-shot scenario, we assume that adapting the LN segmentation foundation model $M_F$ to downstream tasks with a limited number of annotated volumes $\mathcal{D}_d$ can alleviate the target institution's resource constraints. The goal is to fine-tune $M_F$ on a new dataset $\mathcal{D}_d$ with limited labeled data, this can be formulated as:
\begin{equation}
    M_d = \arg\min_{\theta_d \in M_d} \mathcal{L}_{CE+Dice}(\mathcal{D}_d:\theta_d) + R_{penalty}(\theta_F,\theta_d)
    \label{few_shot_finetuning}
\end{equation}
where $M_d$ and $\theta_d$ represent the fine-tuned downstream model and its parameters, respectively; $\mathcal{L}_{CE+Dice}$ is the combined loss; $\mathcal{D}_d$ is the few-shot dataset for the downstream task; The regularization term $R_{penalty}$ helps prevent overfitting and maintain model plasticity by constraining model parameters. This can involve freezing most of the pre-trained parameters $\theta_F$ and updating a small subset (e.g., only fine-tuning the bias~\cite{zaken2022bitfit}) or adding new trainable parameters (e.g., using LoRA~\cite{hulora} or Adapter~\cite{houlsby2019parameter} in key modules).
\subsection{Dynamic Gradient Sparsification Training}
Due to the limited annotations and specific medical image data, advanced data augmentation becomes essential. While more augmentation for training enhances performance, prolonged fine-tuning risks overfitting. Previous work utilized sparse parameter update strategy functions as a regularization mechanism, controlling the upper bound of model stability to mitigate overfitting~\cite{fu2023effectiveness,zhang2024gradient}, but statically sparse the parameter limits the plasticity of the model. To address this, we introduce \textbf{D}ynamic \textbf{G}radient \textbf{S}parsification \textbf{T}raining (DGST), a method designed to achieve a balance between preventing overfitting and maintaining model plasticity during the fine-tuning of UNet-like foundation models under few-shot conditions. In our DGST approach, the gradient is computed at the start of each iteration, specifically at iteration $N$, The gradient set is formulated as: 
\begin{equation}
    G^N = \left\{ g_{\theta_i}^N \bigg| g_{\theta_i}^N = \nabla_{\theta_i} \mathcal{L}(\mathcal{D}_d),\ \theta_i \in \mathcal{P}_O \right\},
    \label{compute gradient}
\end{equation}
where $g^N_{\theta_i}$ is the gradient of parameter $\theta_i$ at iteration $N$; $\mathcal{P}_O$ denotes the set of original model parameters. $\mathcal{L}(\mathcal{D}_d)$ is the loss of the downstream annotated dataset. Further, we assume that the model comprises K convolutional and transposed convolutional kernels, denoted as $\{C_k\}^K_{k=1}$, and select the Top-$\gamma$ gradient parameters with the highest absolute values in each kernel, forming the set of key sparse parameters to be updated, denoted as $\mathcal{P}^N_S$, is formulated as:
\begin{equation}    \mathcal{P}_S^N=\bigcup_{k=1}^K\mathop{\arg\max}\limits_{\theta_i\in\mathcal{P}_O\cap C_k}\limits^{(\gamma)}\left|g_{\theta_i}^N\right|
\label{sparse_parameter_selection}
\end{equation}
where $\mathop{\arg\max}\limits^{(\gamma)}$ refers to get the $\gamma$ parameters with the largest values. Moreover, bias parameters capture task-specific output shifts, while normalization parameters control feature scaling and stability~\cite{frankletraining,zaken2022bitfit}, we include both in $\mathcal{P}_S^N$. Thus, at iteration $N$, the parameter update rule is as follows:
\begin{equation}
\theta_i \leftarrow
\begin{cases}
\theta_i - \eta g_{\theta_i}^N, & \text{for } \theta_i \in \mathcal{P}_s^N \\
\theta_i, & \text{otherwise}
\end{cases}
\end{equation}
where $\eta$ is the learning rate at iteration $N$, $g_{\theta_i}^N$ is calculated in Eq.\ref{compute gradient}. Finally, through iterative execution, the model is enabled to progressively prioritize the most critical parameters. Unlike traditional static sparse parameter constraints~\cite{zhang2024gradient,zaken2022bitfit}, our method implements dynamic sparsity optimization by adjusting parameters via gradient, enabling adaptive updates at each iteration. This approach significantly enhances gradient descent step efficiency, particularly highlighting the effectiveness of our DGST method in achieving a well-balanced trade-off between model flexibility and stability during few-shot fine-tuning.

%
\section{Experiment and Results}
\subsection{Experimental Details}
\textbf{Dataset.} In this study, we employed the RADCURE dataset~\cite{welch2024radcure} to train a foundational model for LN segmentation. An exploratory analysis was performed to delineate 36,106 visible LNs across 3,346 HN CT volumes to assess the foundation model's transferability. Additionally, we also used two publicly available LN segmentation datasets for the few-shot fine-tuning experiment: 120 CT volumes from SegRap2023~\cite{luo2023segrap2023} for HN LN segmentation and 120 CT volumes from LNQ2023~\cite{dorent2025lnq} for mediastinal LN segmentation. Due to the scarcity of labeled data in the few-shot scenario, establishing a stable validation set is challenging. Therefore, we used an 8:2 training-test split. For SegRap2023, few-shot fine-tuning experiments were conducted with 3, 5, and 10 shots, while for LNQ2023, we tested 5, 10, and 20 shots.\\
\begin{table}[t]
  \centering
  \caption{Quantitative comparison of different few-shot fine-tuning methods on two datasets is presented, with results reported as mean ± standard deviation. The best and second-best results are highlighted in bold and underlined, respectively.}
    \resizebox{\textwidth}{!}{
    \begin{tabular}{c|cccccc}
    \toprule
    \multicolumn{7}{c}{\textit{\textbf{SegRap2023}}} \\
    \midrule
    \multirow{2}[2]{*}{Method} & \multicolumn{2}{c|}{\textit{3-shot}} & \multicolumn{2}{c|}{\textit{5-shot}} & \multicolumn{2}{c}{\textit{10-shot}} \\
          & DSC(\%)↑ & \multicolumn{1}{c|}{NSD(\%)↑} & DSC(\%)↑ & \multicolumn{1}{c|}{NSD(\%)↑} & DSC(\%)↑ & NSD(\%)↑ \\
    \midrule
    From scratch & 46.68$\pm$14.46 & \multicolumn{1}{c|}{39.10$\pm$13.86} & 56.52$\pm$9.48 & \multicolumn{1}{c|}{48.15$\pm$7.74} & 58.56$\pm$10.44 & 50.98$\pm$8.30 \\
    Full  & 62.66$\pm$8.45 & \multicolumn{1}{c|}{55.16$\pm$9.86} & 64.86$\pm$9.36 & \multicolumn{1}{c|}{57.68$\pm$9.41} & 65.99$\pm$9.12 & 59.32$\pm$8.21 \\
    LinearProb~\cite{chen2019closer} & 59.31$\pm$11.13 & \multicolumn{1}{c|}{50.48$\pm$10.70} & 58.95$\pm$11.43 & \multicolumn{1}{c|}{50.05$\pm$10.59} & 59.15$\pm$11.37 & 50.31$\pm$10.53 \\
    Bias~\cite{zaken2022bitfit}& 63.48$\pm$9.07 & \multicolumn{1}{c|}{55.04$\pm$9.27} & 66.42$\pm$8.68 & \multicolumn{1}{c|}{58.83$\pm$7.88} & 67.35$\pm$8.72 & 59.96$\pm$7.90 \\
    Adapter~\cite{houlsby2019parameter} & 62.97$\pm$8.92 & \multicolumn{1}{c|}{54.41$\pm$9.74} & \underline{66.77$\pm$8.51} & \multicolumn{1}{c|}{\underline{59.35$\pm$7.69}} & 67.04$\pm$8.80 & 59.69$\pm$8.35 \\
    Lora~\cite{hulora}  & \textbf{65.73$\pm$8.30} & \multicolumn{1}{c|}{\textbf{58.08$\pm$9.24}} & 66.34$\pm$8.59 & \multicolumn{1}{c|}{58.66$\pm$8.06} & 67.02$\pm$8.87 & 59.72$\pm$7.87 \\
    Affine-IN~\cite{frankletraining} & 64.89$\pm$8.44 & \multicolumn{1}{c|}{56.23$\pm$8.47} & 66.33$\pm$8.82 & \multicolumn{1}{c|}{58.75$\pm$7.77} & \underline{67.82$\pm$8.47} & \underline{60.64$\pm$7.50} \\
    \textbf{DGST(Ours)} & \underline{65.05$\pm$8.07} & \multicolumn{1}{c|}{\underline{57.09$\pm$7.18}} & \textbf{67.36$\pm$8.45} & \multicolumn{1}{c|}{\textbf{60.25$\pm$7.62}} & \textbf{68.44$\pm$8.59} & \textbf{61.97$\pm$7.21} \\
    \midrule
    All-shot & \multicolumn{6}{c}{69.20$\pm$9.06, 62.33$\pm$7.47 (96-shot, from scratch)} \\
    \midrule
    \multicolumn{7}{c}{\textit{\textbf{LNQ2023}}} \\
    \midrule
    \multirow{2}[2]{*}{Method} & \multicolumn{2}{c|}{\textit{5-shot}} & \multicolumn{2}{c|}{\textit{10-shot}} & \multicolumn{2}{c}{\textit{20-shot}} \\
          & DSC(\%)↑ & \multicolumn{1}{c|}{NSD(\%)↑} & DSC(\%)↑ & \multicolumn{1}{c|}{NSD(\%)↑} & DSC(\%)↑ & NSD(\%)↑ \\
    \midrule
    From scratch & 37.11$\pm$22.90 & \multicolumn{1}{c|}{34.03$\pm$19.79} & 41.43$\pm$24.18 & \multicolumn{1}{c|}{41.87$\pm$21.12} & 55.76$\pm$25.57 & 55.15$\pm$23.60 \\
    Full  & \underline{49.09$\pm$21.88} & \multicolumn{1}{c|}{\underline{48.72$\pm$19.06}} & \underline{53.68$\pm$24.25} & \multicolumn{1}{c|}{\underline{54.09$\pm$22.06}} & \underline{62.47$\pm$23.44} & \underline{63.00$\pm$21.94} \\
    LinearProb~\cite{chen2019closer} & 6.56$\pm$6.95 & \multicolumn{1}{c|}{6.13$\pm$5.07} & 7.35$\pm$9.08 & \multicolumn{1}{c|}{7.05$\pm$6.84} & 6.88$\pm$7.71 & 6.41$\pm$5.67 \\
    Bias~\cite{zaken2022bitfit}& 44.51$\pm$22.24 & \multicolumn{1}{c|}{40.99$\pm$18.94} & 51.62$\pm$23.68 & \multicolumn{1}{c|}{48.56$\pm$20.82} & 53.43$\pm$21.98 & 50.09$\pm$20.15 \\
    Adapter~\cite{houlsby2019parameter} & 46.99$\pm$23.51 & \multicolumn{1}{c|}{43.43$\pm$20.71} & 51.79$\pm$23.78 & \multicolumn{1}{c|}{48.46$\pm$22.30} & 57.13$\pm$22.48 & 54.48$\pm$20.58 \\
    Lora~\cite{hulora}  & 47.93$\pm$20.72 & \multicolumn{1}{c|}{46.21$\pm$19.06} & 52.13$\pm$22.63 & \multicolumn{1}{c|}{49.78$\pm$20.60} & 59.25$\pm$22.07 & 56.03$\pm$21.62 \\
    Affine-IN~\cite{frankletraining} & 45.61$\pm$21.83 & \multicolumn{1}{c|}{41.75$\pm$19.20} & 51.10$\pm$23.29 & \multicolumn{1}{c|}{47.96$\pm$21.35} & 57.89$\pm$23.08 & 55.16$\pm$21.17 \\
    \textbf{DGST(Ours)} & \textbf{50.94$\pm$22.41} & \multicolumn{1}{c|}{\textbf{48.76$\pm$19.86}} & \textbf{54.94$\pm$24.36} & \multicolumn{1}{c|}{\textbf{55.24$\pm$22.11}} & \textbf{63.82$\pm$21.69} & \textbf{63.84$\pm$20.31} \\
    \midrule
    All-shot & \multicolumn{6}{c}{67.12$\pm$18.65, 67.30$\pm$18.56 (96-shot, from scratch)} \\
    \bottomrule
    \end{tabular}}
  \label{tab:sota}%
\end{table}%
\textbf{Implementation Details.} All training and inference were performed using the nnUNetv2 framework in PyTorch~\cite{isensee2021nnu}, running on a cluster with 8 NVIDIA V100 GPUs. The foundation model was trained and fine-tuned with the default full-resolution 3D U-Net backbone in nnUNet, using a batch size of 2, a patch size of 80$\times$112$\times$224, SGD optimization, an initial learning rate of 0.01 and polynomial decay with power of 0.9 for 2000 epochs. The fine-tuning stage employed the same setup with an initial learning rate of 0.001 for 50 epochs, the hyperparameter $\gamma$ is set to 1. Data augmentation was performed following the default settings of nnUNet. Segmentation performance was assessed in 3D volumes using the Dice similarity coefficient (DSC) for voxel overlap and Normalized Surface Dice (NSD) with a 1 mm tolerance for boundary accuracy~\cite{ma2021abdomenct}.\\
\textbf{Baselines.} We first employed two reference methods: \textbf{From Scratch}, which trains the model from random initialization, and \textbf{Full}, which fine-tunes the entire foundation model. We also employed several PEFT methods: \textbf{LinearProb}~\cite{chen2019closer}, which updates only the classifier head; \textbf{Bias}~\cite{zaken2022bitfit}, which tunes only the bias parameters; and \textbf{Affine-IN}~\cite{frankletraining}, which fine-tunes the affine parameters of the instance normalization layers. Additionally, we applied \textbf{LoRA}~\cite{hulora} and \textbf{Adapter}~\cite{houlsby2019parameter} methods, both of which target only the 3D convolution layers, ensuring efficient fine-tuning through low-rank updates or small auxiliary modules. For comparison, we also used all 96 samples with full-parameter training from scratch, referred to \textbf{All-shot}. All methods were evaluated using the final checkpoint.
\subsection{Results}
\begin{figure}[t]
\includegraphics[width=\textwidth]{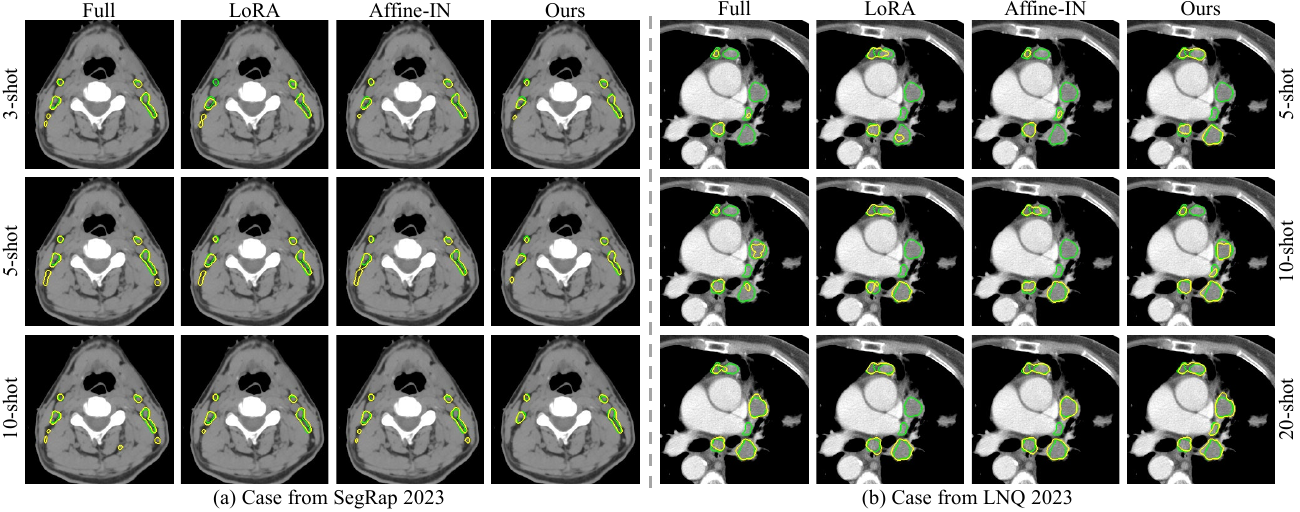}
\caption{Qualitative comparison of different fine-tuning methods. The ground truth and predictions are shown in green and yellow contours, respectively.} \label{sota}
\end{figure}

\begin{figure}[t]
\centering
\includegraphics[width=0.9\textwidth]{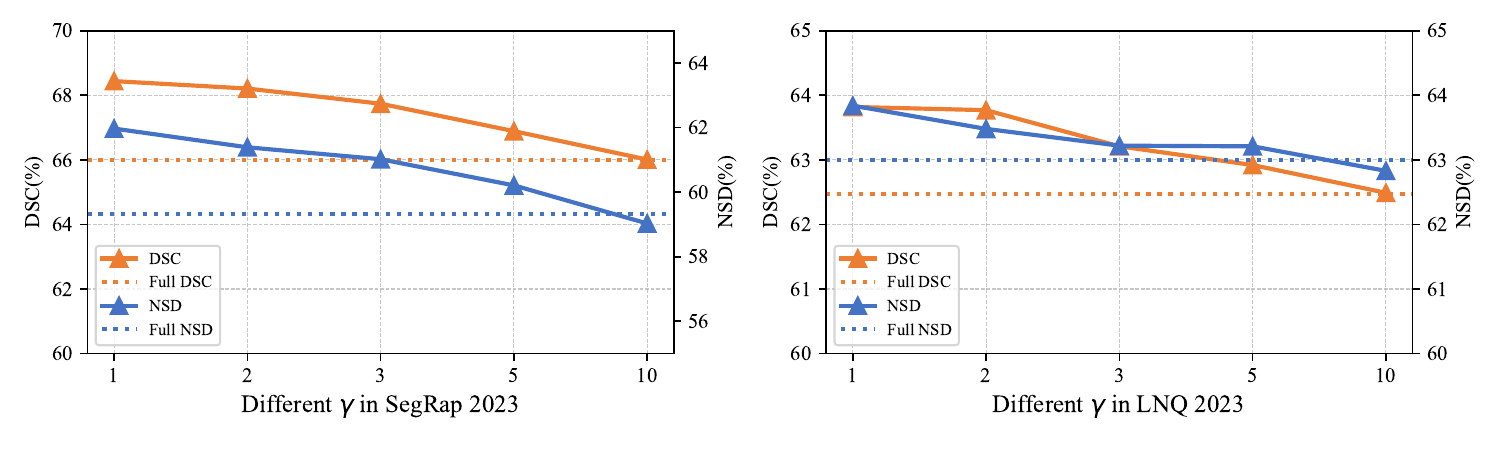}
\caption{Sensitivity analysis of hyperparameter $\gamma$} \label{hyper}
\end{figure}

\textbf{Quantitative and Qualitative Results.} The quantitative performance of different few-shot fine-tuning methods on the SegRap2023 and LNQ2023 datasets is presented in Table~\ref{tab:sota}. For SegRap2023, the "Full" approach achieves better results than "From scratch" with DSC (15.98\%, 8.34\%, 7.43\%) and NSD (16.06\%, 9.53\%, 8.34\%) at 3, 5, and 10-shot, respectively. In LNQ2023, it outperforms DSC (11.98\%, 12.25\%, 6.71\%) and NSD (14.69\%, 12.22\%, 7.85\%) in 5, 10 and 20-shot, respectively. These results demonstrate our LN segmentation foundation model's strong transferability in both HN and mediastinal areas. In the SegRap2023 dataset, which has a lower domain discrepancy, all few-shot fine-tuning methods, except for "LinearProb", outperform the "Full" method, highlighting the advantage of the sparse parameter strategy in mitigating overfitting. Our DGST method achieved the second-best result in the 3-shot setting and the best performance in the 5-shot and 10-shot settings. These results suggest that, with slightly more training samples, our method effectively balances anti-overfitting and plasticity, demonstrating superior flexibility. In contrast, for the higher-domain gap dataset LNQ2023, other methods fail to outperform the "Full" approach, whereas our method consistently delivers the best results across all settings. Fig.\ref{sota} shows qualitative results between different methods in both two downstream tasks. While all methods scale with training data, only our DGST consistently excels in reducing both false positives and missed detections for LN segmentation. Both quantitative and qualitative results demonstrate DGST's capability to maintain model plasticity, enabling continuous learning of intricate LN patterns across diverse clinical scenarios.\\
\textbf{Sensitivity Analysis of $\gamma$.} We examined the influence of the hyperparameter $\gamma$ in the DGST method, which controls the selection of top gradient parameters for sparsification. Experiments were conducted on the SegRap2023 (10-shot) and LNQ2023 (20-shot) datasets with $\gamma$ values of 1, 2, 3, 5, and 10. Additionally, we present the results of the "Full" method for comparison. As shown in Fig.\ref{hyper}, the results indicate that smaller $\gamma$ values produce similar performance, while larger values lead to a performance decline, falling close to the "Full".\\
\textbf{Ablation study.} We performed an ablation study to evaluate different parameter sparsification strategies, with the baseline "Full" method as a reference. We first fine-tuned only the encoder ("Encoder Only") or decoder ("Decoder Only"), finding that sparsification on general structures underperformed. We then explored tuning bias and normalization parameters ("Bias+Norm") and tested a modified approach with random parameter selection ("Dynamic Random Sparsification Training, DRST"). While sparsification improved SegRap2023 performance compared to the "Full" method, it failed to enhance results on LNQ2023, suggesting insufficient consideration of model plasticity. We further implemented a static parameter selection approach ("Static Gradient Sparsification Training, SGST") via initial accumulated gradients. Comparisons of performance and time cost for each strategy are shown in Table \ref{tab:ablation}, demonstrating that gradient-based parameter selection is crucial for model plasticity. However, effective sparsification necessitates dynamic updates based on current gradients to balance stability and flexibility, albeit at the cost of micro-longer times.
\begin{table}[t]
  \centering
  \caption{Ablation study on different parameter sparsification strategies for the few-shot fine-tuning of our LN segmentation foundation model. Full: Full parameters fine-tuning; Bias+Norm: Tuning on bias and normalization parameters; DRST: Dynamic Random Sparsification Training; SGST: Static Gradient Sparsification Training.}
  \resizebox{\textwidth}{!}{
    \begin{tabular}{c|cccc|cccc|c}
    \toprule
  \multirow{3}[4]{*}{Method} & \multicolumn{4}{c|}{\textit{\textbf{SegRap2023}}} & \multicolumn{4}{c|}{\textit{\textbf{LNQ2023}}} & \multirow{3}[4]{*}{\makecell{Iteration\\Duration}} \\
\cmidrule{2-9}          & \multicolumn{2}{c}{\textit{3-shot}} & \multicolumn{2}{c|}{\textit{10-shot}} & \multicolumn{2}{c}{\textit{5-shot}} & \multicolumn{2}{c|}{\textit{20-shot}} &  \\
          & DSC(\%) & NSD(\%) & DSC(\%) & NSD(\%) & DSC(\%) & NSD(\%) & DSC(\%) & NSD(\%) &  \\
    \midrule
    Full   & 62.66$\pm$8.45 & 55.16$\pm$9.86 & 65.99$\pm$9.12 & 59.32$\pm$8.21 & 49.09$\pm$21.88 & 48.72$\pm$19.06 & 62.47$\pm$23.44 & 63.00$\pm$21.94 & 0.2085s \\
    Encoder Only & 59.33$\pm$11.13 & 50.48$\pm$10.69 & 59.23$\pm$11.34 & 50.37$\pm$10.53 & 6.49$\pm$6.86 & 6.03$\pm$5.02 & 6.92$\pm$7.72 & 6.43$\pm$5.66 & 0.0880s \\
    Decoder Only & 59.08$\pm$10.81 & 51.12$\pm$11.28 & 64.24$\pm$9.96 & 57.37$\pm$8.45 & 46.97$\pm$24.15 & 45.21$\pm$21.52 & 59.68$\pm$22.24 & 58.88$\pm$20.17 & 0.1605s \\
    Bias+Norm & 64.21$\pm$9.08 & 55.79$\pm$9.17 & 67.26$\pm$8.24 & 60.21$\pm$7.63 & 45.91$\pm$21.27 & 41.82$\pm$19.34 & 56.12$\pm$21.32 & 54.19$\pm$20.13 & 0.1534s \\
     DRST & \textbf{65.61$\pm$8.53} & \textbf{57.64$\pm$8.30} & 67.59$\pm$8.93 & 60.20$\pm$7.83 & 43.56$\pm$21.12 & 40.44$\pm$18.05 & 52.19$\pm$22.13 & 48.93$\pm$20.91 & 0.2193s \\
    SGST & 64.58$\pm$7.99 & 56.84$\pm$8.40 & 67.78$\pm$8.65 & 61.01$\pm$6.89 & 48.25$\pm$21.40 & 46.44$\pm$19.01 & 60.28$\pm$25.18 & 60.02$\pm$23.70 & 0.2197s \\
    \textbf{DGST} & 65.05$\pm$8.07 & 57.09$\pm$7.18 & \textbf{68.44$\pm$8.59} & \textbf{61.97$\pm$7.21} & \textbf{50.94$\pm$22.41} & \textbf{48.76$\pm$19.86} & \textbf{63.82$\pm$21.69} & \textbf{63.84$\pm$20.31} & 0.2813s \\
    \bottomrule
    \end{tabular}}
  \label{tab:ablation}%
\end{table}%
\section{Conclusion}
In conclusion, this study introduces DGST, a novel approach designed to enhance the fine-tuning of foundation models for LN segmentation with few-shot annotations. By dynamically selecting and updating the most critical parameters based on the gradient at each iteration, DGST effectively balances model stability and flexibility, mitigating the risks of overfitting while preserving the model's ability to adapt to new medical scenarios. The results across SegRap2023 and LNQ2023 demonstrate the superior performance of DGST compared to existing fine-tuning methods. Besides, we will release the dataset of 36,106 annotated LNs and the validated framework to advance the deployment of robust, resource-efficient segmentation tools in evolving clinical workflows.

\subsubsection{\ackname}
This work was supported by the National Natural Science Foundation of China under Grant 82203197 and 62271115, the Sichuan Science and Technology Program, China (Grant 2023NSFSC1852 and 2023NSFSC0720), and the Sichuan Provincial Cadre  Health Research Project (Grant/award number: 2023-803). We would like to thank Department of Radiation Oncology, the Sichuan Cancer Hospital for the data annotation and checking. 
\subsubsection{\discintname}
The authors have no competing interests to declare that are relevant to the content of this article.
%
%
%
%
\bibliographystyle{splncs04}
\bibliography{Lymph.bib}
\end{document}